\documentclass[letterpaper]{article} 
\usepackage{aaai2027}  
\nocopyright
\usepackage[hyphens]{url}  
\usepackage{graphicx} 
\usepackage{natbib}  
\usepackage{caption} 
\usepackage{algorithm}
\usepackage{algorithmic}

\usepackage{newfloat}
\usepackage{listings}
\DeclareCaptionStyle{ruled}{labelfont=normalfont,labelsep=colon,strut=off} 
\floatstyle{ruled}
\newfloat{listing}{tb}{lst}{}
\floatname{listing}{Listing}

\usepackage{booktabs}

\title{Contiguity, Not Importance: Budgeted Repair of Stale KV Caches After Document Edits}

\author{
    Mingyang Mao\textsuperscript{\rm 1},
    Wyatt Mackey\textsuperscript{\rm 2},
    Xiaomin Lin\textsuperscript{\rm 1}\corresponding
}

\affiliations{
    \textsuperscript{\rm 1}University of South Florida, Electrical and Computer Engineering\\
    \textsuperscript{\rm 2}DEVCOM Army Research Laboratory\\
    \{mmao, xlin2\}@usf.edu, wyatt.t.mackey.civ@army.mil
}

\begin{document}

\maketitle

\begin{abstract}
KV-cache reuse can reduce inference cost in retrieval-augmented generation and agentic systems, but cached contexts may become stale when retrieved knowledge, working memory, or user state is edited. Under causal self-attention, even a local edit can affect downstream KV states. A full re-prefill reliably restores consistency but is costly, whereas refreshing only the edited span can leave downstream dependencies stale. We formulate in-place repair as budgeted recomputation and compare training-free position-selection policies on a factual RAG benchmark with matched direct and derived edits. Across three model families, all policies repair direct cases, but derived cases clearly separate them. At the primary budget, a contiguous edit-local window recovers at least 0.94 of the post-edit answer margin and substantially outperforms attention-based, KV-deviation, and structural selectors. Mechanistic analysis shows that position sets effective under clean-state transplantation can fail under actual recomputation because scattered positions inherit surrounding staleness. The edit-local advantage also depends on adjacency and largely disappears when the answer-bearing text moves downstream. Because answer-relevant edits almost always corrupt model behavior, failure severity is difficult to predict, and repair is 13–21× faster than full re-prefill, our results support unconditional edit-local repair when the dependent text remains adjacent to the edit.

\end{abstract}


\section{Introduction}

\begin{figure}[!t]
    \centering
    \includegraphics[width=0.8\linewidth]{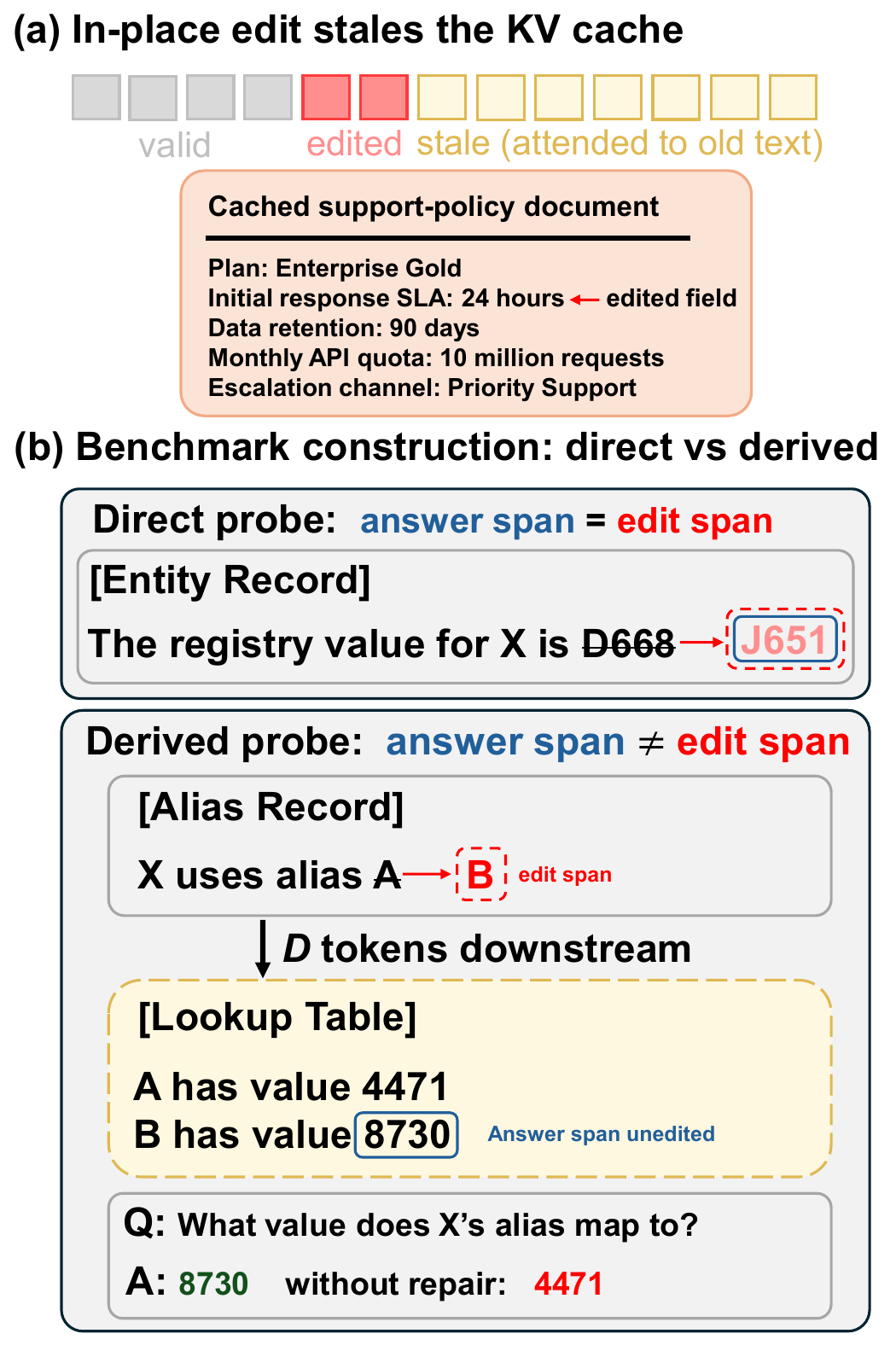}
    \caption{\textbf{Problem and probe design.} (a)~An in-place edit to a cached
document leaves the edited span's KV entries encoding old text and all
downstream entries stale. (b)~Each benchmark item pairs the same question with
two context structures: a \emph{direct} probe, where the edit rewrites the
answer text, and a \emph{derived} probe, where the
answer sits in an unedited lookup entry downstream of the edited alias. Without repair the cache returns the pre-edit answer.}
    \label{fig:figure1}
\end{figure}



RAG pipelines and LLM-based agents increasingly reuse KV states for
retrieved documents or memory blocks across requests, avoiding repeated prefill
\cite{yao2025cacheblend,bergman2025leveraging,ye2026kvcomm,pan2026kvflow}.
This optimization assumes cached source content does not change. In
deployment, that assumption breaks when a factual record is corrected
\cite{ouyang2025hoh,cohen2024ripple}, a policy is revised, or an agent's
working memory is updated \cite{packer2023memgpt}. The source text then carries the update, while retained
KV tensors encode the older version. A request can therefore be answered from a
representation that no longer matches the knowledge base. Existing
document-level reuse methods optimize reuse and composition, but do not
synchronize cached representations after a source edit.


Repairing this mismatch is not confined to the edited span. Continuing to
serve from a stale cache can reduce response accuracy
\cite{ouyang2025hoh}, and causal self-attention leaves every downstream KV
state dependent on the old content. A full re-prefill restores consistency
but re-encodes an entire long context after an edit of only a few tokens.
Recent work has begun to repair the cache in place after such small edits, but each proposed method carries a limitation. KVEraser replaces a target span with learned steering states, yet requires per-model training and targets deletion rather than factual replacement\cite{li2026kveraser}. MTN recomputes downstream \textit{notes}
ranked by their causal effect, using an oracle signal in controlled agent
tasks \cite{li2026MTN}. Neither answers the matched-budget question for
factual RAG under actual recomputation. To build a low-budget, training-free repair policy, we therefore first ask which downstream positions matter most once the edited span itself has been refreshed.

Our study pairs factual edits with contexts of roughly
$5000$ retrieved tokens. Each question either asks for the edited fact
directly or requires a two-hop derivation through unchanged downstream text.
Starting from the stale cache, every policy refreshes the known edit span and
selects $K$ downstream positions under the same recomputation budget. We
compare an edit-local window, structural delimiters \cite{li2026MTN},
stale-query attention \cite{wang2026prophetkv}, CacheBlend-based KV deviation
\cite{yao2025cacheblend}, and random selection. A transplant-derived causal
ranking is included only as a non-deployable diagnostic. Evaluation covers
development and held-out cohorts across Llama, Qwen, and Mistral, plus a
$1099$-item variant that moves the answer-carrying text downstream. We use
pre-specified criteria and report held-out margin recovery and flip rate.



Three results define the baseline. At $K{=}32$, every policy saturates
on direct questions, while derived answers separate them. The edit-local
window recovers $0.94$--$1.01$ of the answer margin and flips
$0.95$--$1.00$ of held-out items, beating every alternative by
$0.46$--$1.00$ margin recovery (Holm-corrected
$p \le 3\times10^{-9}$). This advantage depends on adjacency. Moving the
answer-carrying text $250$ tokens downstream reduces edit-local recovery to
$0.01$--$0.09$, with stale-query attention helping on only one model family.
Transplant recoverability also fails to predict repair under recomputation.
The causal ranking recovers $0.92$--$0.97$ of the margin when clean states are
transplanted, but the same positions recover only $0.01$--$0.21$ when
recomputed at $K{=}8$. Recomputation is worse on every held-out item. A
transplanted state imports information from a correct cache, whereas a
recomputed state reads stale surrounding states. A contiguous edit-local
window succeeds by rebuilding that dependency chain in order.
Finally, answer-relevant edits almost always break the stale cache
(base rate $\ge 0.988$), and cheap features poorly predict failure severity
(best out-of-fold $\rho=0.17$). Because repair runs $13$--$21\times$ faster
than re-prefill, applying edit-local repair unconditionally is the practical
policy within this setting. It is also the training-free baseline that future
selection methods must beat.

In summary, this paper makes the following key contributions:

\begin{itemize}
    \item We formulate stale-cache repair as budgeted recomputation and
    introduce paired direct, derived, and distance-controlled factual edits.
    \item We compare five training-free selectors at a matched budget across
    three model families, establishing \textsc{EditLocal} as a strong
    adjacent-block baseline.
    \item We separate localization under clean-state transplant from repair
    under recomputation.
    \item We show that answer-relevant edits warrant unconditional repair
    within this benchmark because failure is frequent, severity is hard to
    predict, and repair is inexpensive.
\end{itemize}


\section{Related Works}

\paragraph{Document-level KV reuse.}
RAGCache stores intermediate states of retrieved knowledge, while TurboRAG
precomputes per-chunk KV caches for reuse across queries
\cite{jin2025ragcache,lu2025turborag}. Both reduce repeated prefill by reusing
stored states of retrieved text. Neither addresses how those states should be
updated after the source text changes. HoH shows that outdated retrieved
evidence can reduce RAG accuracy even when current evidence is available
\cite{ouyang2025hoh}, but it studies stale information at the text level rather
than repair of an already cached representation. We study the systems problem
that follows a source edit: how much of one stale document cache must be
recomputed?

\paragraph{Selective recomputation.}
Selective recomputation addresses a different cache mismatch. CacheBlend
restores missing cross-attention by recomputing positions with large KV
deviations, while ProphetKV prioritizes positions using query relevance
\cite{yao2025cacheblend,wang2026prophetkv}. EPIC recomputes a small, fixed set
of initial chunk tokens, Cache-Craft repairs reusable chunk caches through
limited recomputation, and KVShare selects high-deviation states during prefill
and decoding \cite{hu2025epic,agarwal2025cachecraft,yang2025kvshare}. Across
these methods, the source chunks are unchanged and the mismatch comes from
reusing their caches in a new context. Our experiments transfer the deviation-
and query-based selection signals to a cache whose source text has changed and
compare them at the same downstream-position budget.

\paragraph{KV-cache editing.}
KV-cache editing is closest to our setting. KVEraser replaces a target span
with learned steering states to erase its influence \cite{li2026kveraser}.
\citeauthor{li2026MTN} \shortcite{li2026MTN} shows that changing a field can
leave old conclusions in downstream states and ranks those states by causal
effect. Leyline provides serving primitives for removing or replacing cached
spans, including positional correction for length-changing edits
\cite{ma2026leyline}. We isolate length-preserving factual replacement and ask
which training-free selector works when its chosen positions are actually
recomputed at a matched budget. This separates our setting from learned
erasure, cache splicing, and position sets ranked through oracle-state
transplantation. 

%
%

\section{Cache Repair as Budgeted Recomputation}

\paragraph{Question Setting.}
A serving system prefills a context $C$ and stores its key--value states for
reuse. After an edit produces $C'$, the stored cache
$\mathcal{K}(C)$ is stale; a full prefill $\mathcal{K}(C')$ is the
\emph{oracle} repair target, not a competing method. In-place repair instead
refreshes selected entries of $\mathcal{K}(C)$ toward this target.

We isolate one contiguous, length-preserving edit:
$|C|=|C'|=n$ and $c_i=c'_i$ outside the half-open span
$S=[s_{\mathrm{start}},s_{\mathrm{end}})$. Positions and rotary phases
therefore remain fixed; length-changing edits, which also require positional
correction, are outside our scope \cite{ma2026leyline}. Each benchmark item
pairs a \emph{direct} condition, whose answer lies in $S$, with a
\emph{derived} condition, whose answer lies in unchanged downstream text but
depends on the edited fact (Figure~\ref{fig:figure1}b). The pair otherwise
shares its retrieved documents, insertion point, subject, value pair, and
query, and we report the conditions separately.

\paragraph{The repair interface.}
Every policy uses the same operator and differs only in its $K$ selected
positions. Let $D=[s_{\mathrm{end}},n-1)$ be the downstream candidate pool
and let $A_\pi\subseteq D$, $|A_\pi|=K$, be the positions chosen by policy
$\pi$. The repaired set is $P_\pi=S\cup A_\pi$, and
\[
\widetilde{\mathcal{K}}
  = \mathcal{R}\big(\mathcal{K}(C),C',P_\pi\big).
\]
The operator recomputes $P_\pi$ from $C'$ layer by layer and leaves all other
states unchanged, using the same selective-recomputation primitive as prior
systems \cite{yao2025cacheblend}. Its full-position endpoint is
\begin{equation}
\mathcal{R}\big(\mathcal{K}(C),\, C',\, \{0,\dots,n-1\}\big)
  \;=\; \mathcal{K}(C') .
\label{eq:endpoint}
\end{equation}

The known edit span is always recomputed and shared across policies, while
the fresh query is never cached and is recomputed during scoring. Thus, $K$
counts only additional downstream positions. We exclude upstream positions
because causal attention prevents them from depending on the edit, and verify
this invariant numerically. The question is which deployable rule chooses
$A_\pi$ most effectively.

\paragraph{Cost accounting.}
We separate three costs: common work for the edit span and fresh query, the
matched repair budget of $K\cdot L$ token-layers, and selector overhead.
The common term cancels in policy comparisons, while overhead is reported
outside $K$. Text- and position-only rules have no selector forward;
\textsc{Attention} adds a stale-cache scoring pass
\cite{wang2026prophetkv}, and \textsc{CacheBlend} adds roughly one prefill
layer over the prefix \cite{yao2025cacheblend}. We use wall-clock latency as
the headline systems measure and token-layers as a device-independent
secondary measure; Section~\ref{sec:results} reports both. The full
construction, interface invariants, and cost decomposition appear in
the supplementary appendix.

\paragraph{What counts as repaired.}
Answer accuracy alone cannot distinguish caches that emit the same string
with different underlying preferences. Each item therefore defines a
pre-edit answer $a_{\mathrm{old}}$ and a post-edit answer
$a_{\mathrm{new}}$. For cache state $x$, let $m_x$ be the difference between
their length-averaged, teacher-forced log probabilities. Our primary metric is
\begin{equation}
\mathrm{MR} \;=\;
\frac{m_{\mathrm{repaired}} - m_{\mathrm{stale}}}
     {m_{\mathrm{oracle}} - m_{\mathrm{stale}}},
\label{eq:mr}
\end{equation}
where $0$ denotes no change from the stale cache and $1$ reaches the oracle.
MR is unclipped, and zero-gap items are excluded.

We pair this internal measure with visible behavior. Let $y_b$ be the greedy
continuation of at most $16$ tokens for retained item $b$. The flip rate is
\begin{equation}
\mathrm{Flip} \;=\; \frac{1}{|\mathcal{B}|}\sum_{b \in \mathcal{B}}
\mathbf{1}\big[\,a_{\mathrm{new}} \in y_b \;\wedge\;
a_{\mathrm{old}} \notin y_b\,\big].
\label{eq:flip}
\end{equation}
MR captures partial recovery before the decision boundary, whereas flip
records whether generation exposes only the new answer. We report both, with
KL recovery as a secondary distributional measure. The appendix
gives the complete scoring and edge-case rules.

\paragraph{Selection policies.}
Table~\ref{tab:policies} compares four deployable signals---edit proximity,
structure, stale-query attention \cite{wang2026prophetkv}, and KV deviation
\cite{yao2025cacheblend}---with a matched random control. These policies may
use the stale cache, edited context, edit location, and query, but never the
oracle cache or answers. Grey rows use unavailable oracle or carrier
information and are diagnostics, not deployable methods or upper bounds on
recomputation.

\begin{table}[t]
\centering
\small
\begin{tabular}{@{}lp{0.62\columnwidth}@{}}
\toprule
Policy & Rule (all pick $K$ positions from $[s_{\mathrm{end}}, n{-}1)$) \\
\midrule
\textsc{EditLocal} & the $K$ positions immediately after the span \\
\textsc{Structural} & structural tokens, nearest to the span first \\
\textsc{Attention} & top-$K$ by stale-cache attention from the query \\
\textsc{CacheBlend} & top-$K$ by layer-2 key/value deviation \\
\textsc{Random} & uniform from the pool, $5$ seeds \\
\midrule
\textcolor{gray}{\textsc{CausalOracle}} &
  \textcolor{gray}{top-$K$ by measured causal effect (needs oracle cache)} \\
\textcolor{gray}{\textsc{CarrierWindow}} &
  \textcolor{gray}{contiguous window at the true carrier block (needs its
  location)} \\
\bottomrule
\end{tabular}
\caption{Selection policies, frozen before held-out evaluation. Only $K$
varies; grey rows are diagnostics.}
\label{tab:policies}
\end{table}

The primary budget is $K{=}32$; curves over
$K\in\{8,16,32,64\}$ are descriptive. The structural
\textsc{EditLocal}$@\partial$ variant in Section~\ref{sec:extent} is
exploratory because it was inferred from those curves. Full policy definitions
and provenance appear in the appendix.

\paragraph{Transplant is not recomputation.}
A selected position set can be used in two ways. Transplant copies its KV
states from the oracle cache into the stale cache, while recomputation rebuilds
them within the stale cache. Only recomputation is deployable because
transplant requires the oracle cache. We use transplant only as a localization
diagnostic.

The two operators provide different information. A transplanted state was
computed with an otherwise correct cache and can therefore import information
from outside the selected set. A recomputed state must instead read the
still-stale cache around it. Positions that appear sufficient under transplant
may therefore fail under recomputation. We apply both operators to identical
position sets to measure this gap.

\begin{figure*}[t]
\centering
\includegraphics[width=\textwidth]{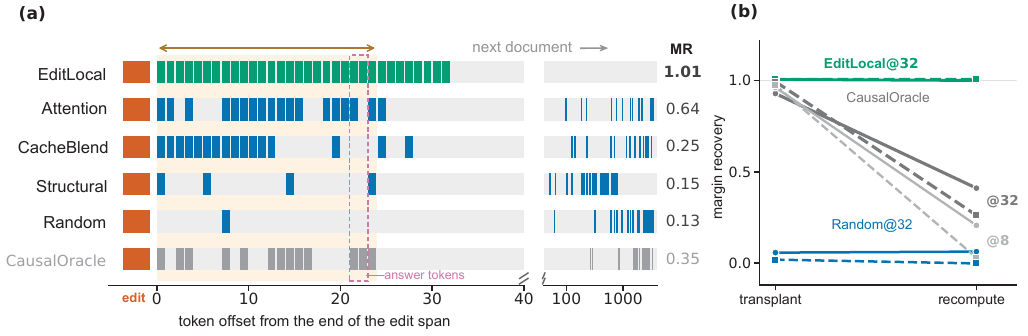}
\caption{\textbf{Contiguity, not localization, decides repair.}
(a)~Selections for a representative held-out Llama item at $K{=}32$.
Orange marks the edit span, repaired outside the budget, and grey marks
states left stale. Only \textsc{EditLocal} forms an unbroken path from
the edit through the answer. (b)~Margin recovery under transplant and
recomputation on identical sets from $15$ development items. Lines
connect the two operators. Solid lines denote Llama and dashed lines
Qwen. Grey \textsc{CausalOracle} results are mechanism diagnostics
without tests.}
\label{fig:budget}
\end{figure*}

\section{Experimental Setup}

\paragraph{Paired edit benchmark.}

Each \emph{item} contains roughly $5000$ tokens of shuffled HotpotQA
\texttt{fullwiki} validation paragraphs \cite{yang2018hotpotqa}, used only as
realistic retrieval background; HotpotQA questions and answers are not used. We insert one
synthetic record at a document boundary about $30\%$ into the context.
\emph{Background} denotes the filler paragraphs, whereas \emph{item} denotes
the full construction and is the unit of every paired comparison. The direct
block states the queried field and rewrites its value. The derived block
instead rewrites an alias that selects between two downstream lookup rows;
both rows are present before the edit and remain unchanged. Thus, only the
location of the answer relative to the edit differs between the conditions.
All synthetic subjects, aliases, and values are absent from the background.

Build-time filters require the old and edited prefixes to have equal length
under every tokenizer, differ in one span of at most six tokens, and place
that span between $25\%$ and $40\%$ of the context. Direct and derived pairs
must match within $3\%$ in relative edit position. At evaluation, we retain
only items for which the oracle answers correctly and a stale-to-oracle margin
gap exists. These model-dependent filters define retention.
The appendix gives the full construction and filter provenance.

\paragraph{Distance-controlled variant.}
The main derived block makes the edit and answer-carrying text adjacent. To
vary this geometry, we split its alias and lookup records into two documents,
leave the alias in place, and move the lookup record downstream. We define
$d=s_{\mathrm{carrier}}-s_{\mathrm{end}}\in\{0,250,1500\}$. The direct
condition appears only at $d{=}0$. At $d{=}0$, the derived construction is
byte-identical to the main benchmark. The distance axis extends rather than
replaces the main setting; all policies, budgets, and metrics remain fixed.

\paragraph{Models and decoding.}
We use Llama-3.1-8B-Instruct, Qwen3-8B, and Mistral-7B-Instruct-v0.3 in
bfloat16 with SDPA attention, decoding greedily with no chain of thought.
Qwen3 receives an empty think block so that it answers directly. Because the
three tokenizers map the same text to different token counts, a fixed $K$
does not cover identical text spans across families. Before evaluation, a
frozen harness gate compares the repaired cached path with a full forward at
the answer position under a bfloat16 tolerance fixed after the initial smoke
test.

\paragraph{Pre-specified held-out evaluation.}
Development uses $100$ items. The policy roster, metrics, budgets, and tests
are pre-specified and applied consistently across all three model families.
Held-out items use different seeds and HotpotQA indices disjoint from
development. A candidate-pool expansion rule is fixed in advance, preventing
retention shortfalls from being repaired by loosening filters. Retention is
$75$ direct and $34$ derived for Llama, $75$ in each condition for Qwen, and
$60$ direct and $58$ derived for Mistral. Llama's lower derived count reflects
failure to solve some two-hop items even from a clean prefill.

All comparisons are paired by item. We report $95\%$ percentile intervals
from $10{,}000$ bootstrap resamples. A policy mean is resampled directly;
a contrast is formed per item before resampling so that pairing is preserved.
Paired Wilcoxon tests accompany the contrasts. At $K{=}32$, all $15$
comparisons among six policies form one Holm-corrected family within each
model and condition; \textsc{CarrierWindow} appears only in the distance
variant. Curves over $K$ are descriptive and are not tested. Win--tie--loss
uses a primary MR tie band of $0.05$, fixed before held-out evaluation from
the spread of semantically equivalent arms, with $0.01$ and $0.10$ as
appendix sensitivities. Direct and derived conditions are never pooled.
The appendix records the cohort and statistical
details.

\section{Results}

\label{sec:results}

Across all three model families, every policy saturates in the direct
condition, which serves as a sanity check. The derived condition
separates the methods: \textsc{EditLocal} nearly restores the full
answer margin when the dependent text is adjacent to the edit, but its
advantage largely disappears once that text moves downstream.
Transplant-ranked positions also lose most of their value under actual
recomputation. Together with near-universal stale-cache failure and the
low cost of repair, these results support unconditional edit-local
repair within the adjacent-block setting.

\begin{table*}[t]
\centering
\small
\begin{tabular}{@{}lrcrcrc@{}}
\toprule
& \multicolumn{2}{c}{Llama-3.1-8B ($n{=}34$)} & \multicolumn{2}{c}{Qwen3-8B ($n{=}75$)} & \multicolumn{2}{c}{Mistral-7B ($n{=}58$)} \\
\cmidrule(lr){2-3}\cmidrule(lr){4-5}\cmidrule(l){6-7}
Policy & MR & Flip & MR & Flip & MR & Flip \\
\midrule
\textsc{EditLocal} & \textbf{0.993} $\pm$ 0.008 & 0.97 & \textbf{1.007} $\pm$ 0.005 & 1.00 & \textbf{0.937} $\pm$ 0.032 & 0.95 \\
\textsc{Structural} & 0.126 $\pm$ 0.065 & 0.03 & 0.008 $\pm$ 0.011 & 0.01 & 0.012 $\pm$ 0.012 & 0.00 \\
\textsc{Attention} & 0.533 $\pm$ 0.068 & 0.03 & 0.015 $\pm$ 0.014 & 0.01 & 0.019 $\pm$ 0.014 & 0.00 \\
\textsc{CacheBlend} & 0.238 $\pm$ 0.080 & 0.03 & 0.076 $\pm$ 0.030 & 0.03 & 0.012 $\pm$ 0.012 & 0.00 \\
\textsc{Random} & 0.078 $\pm$ 0.054 & 0.00 & 0.004 $\pm$ 0.010 & 0.00 & 0.004 $\pm$ 0.009 & 0.00 \\
\midrule
\textcolor{gray}{\textsc{CausalOracle}} & \textcolor{gray}{0.409 $\pm$ 0.064} & \textcolor{gray}{0.00} & \textcolor{gray}{0.187 $\pm$ 0.037} & \textcolor{gray}{0.09} & \textcolor{gray}{0.096 $\pm$ 0.035} & \textcolor{gray}{0.05} \\
\bottomrule
\end{tabular}
\caption{Held-out results for the derived condition at $K{=}32$.
MR is Eq.~\ref{eq:mr}, reported as the mean $\pm$ the half-width of a
$95\%$ percentile interval from $10{,}000$ bootstrap resamples, which
is symmetric to within $0.003$; Flip is Eq.~\ref{eq:flip}. Every paired
contrast against \textsc{EditLocal} is Holm-significant within its
model's $15$-pair family ($p \le 3\times10^{-9}$). The grey row is a
transplant-ranked diagnostic, not a method or an upper bound under
recomputation. Direct rows are omitted because all policies achieve
$0.98$--$1.00$.}
\label{tab:main}
\end{table*}

\subsection{The Edit-Local Window Wins}

At the frozen $K{=}32$, \textsc{EditLocal} recovers $0.993$, $1.007$,
and $0.937$ of the margin on Llama, Qwen, and Mistral and flips
$95$--$100\%$ of held-out items (Table~\ref{tab:main}). The best rival,
\textsc{Attention} on Llama, reaches $0.533$. No deployable rival clears
$0.08$ on Qwen or Mistral. The grey \textsc{CausalOracle} reaches only
$0.10$--$0.41$. On Llama, rivals can move probability toward the new
answer, but none flips more than one of $34$ items. MR thus captures
partial internal repair, while Flip records whether generation exposes
the new value.

Figure~\ref{fig:budget}a makes the geometric difference visible.
\textsc{CausalOracle} includes the answer token but recovers only
$0.35$, and \textsc{CacheBlend} stops inside the row holding the old
value. Only \textsc{EditLocal} rebuilds an unbroken path from the edit
through the dependent text. It beats the oracle on all $34$, $75$, and
$58$ held-out items, and its weakest record against any rival is
$32$ wins, $2$ ties, and no losses. KL recovery follows MR at $0.999$,
$1.000$, and $0.943$.

Qwen's mean MR above one reflects slight overshoot beyond the oracle
margin, which Eq.~\ref{eq:mr} leaves visible. Mistral reaches $1.002$
at $K{=}64$, consistent with its tokenizer requiring more than
$32$ tokens to cover the same text.

\begin{figure*}[t]
\centering
\includegraphics[width=\textwidth]{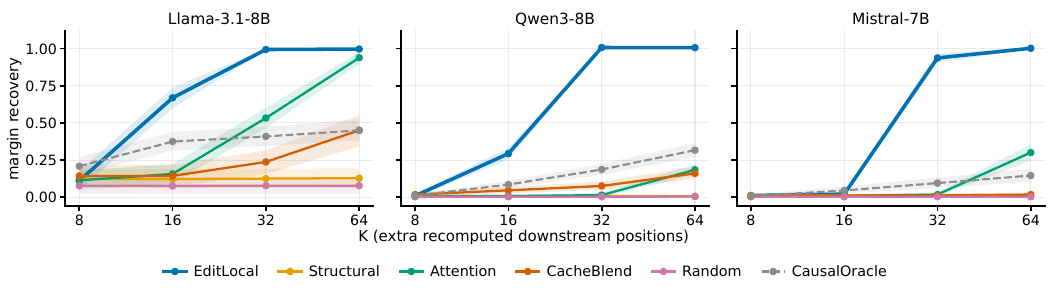}
\caption{Margin recovery against the budget $K$ on the held-out set,
derived condition. Budgets double along a logarithmic axis, and bands
are $95\%$ bootstrap intervals. The direct condition is omitted because every
policy lies between $0.98$ and $1.00$ from $K{=}8$ on. The curves are
descriptive, and the specification reserves testing for $K{=}32$.
\textsc{EditLocal} is threshold-shaped and saturates once the window
covers the remainder of the injected block. The grey dashed line is the
transplant-ranked oracle, a diagnostic rather than a method.}
\label{fig:kcurves}
\end{figure*}

\subsection{How Far the Window Should Extend}
\label{sec:extent}

Direct results stay near one at every budget and are omitted from
Figure~\ref{fig:kcurves}. In the derived condition,
\textsc{EditLocal} rises from $0.11$ to $1.00$ on Llama, from $0.01$
to $1.01$ on Qwen, and from $0.01$ to $1.00$ on Mistral as
$K$ increases from $8$ to $64$. The transition occurs at $K{=}32$.
At $K{=}64$, \textsc{Attention} reaches $0.94$ on Llama but only
$0.18$ and $0.30$ on Qwen and Mistral, and every other rule remains at or
below $0.45$. More budget alone does not produce the same gain.

The threshold matches the block geometry, an exploratory observation
made after unblinding the curves. From the edit to the block boundary,
the dependent text spans a median of $24$ Llama tokens (maximum $26$)
and $28$ Qwen tokens (maximum $30$). A window of $16$ tokens is
incomplete, while $32$ covers the path. This motivates
\textsc{EditLocal}$@\partial$, which recomputes to the next document
boundary with no selector forward. It coincides with $K{=}32$ on Llama
and Qwen, while Mistral needs a slightly larger budget. This structural rule
is an interpretation of the curves, not a separately tested arm, and
still assumes that dependent text begins where the edit ends.

\begin{table*}[t]
\centering
\small
\begin{tabular}{@{}lrrrrrrrr@{}}
\toprule
& \multicolumn{2}{c}{Llama} & \multicolumn{3}{c}{Qwen} & \multicolumn{3}{c}{Mistral} \\
\cmidrule(lr){2-3}\cmidrule(lr){4-6}\cmidrule(l){7-9}
Policy & $d{=}0$ & $250$ & $d{=}0$ & $250$ & $1500$ & $d{=}0$ & $250$ & $1500$ \\
\midrule
\textsc{EditLocal} & 1.00 & 0.09 & 1.00 & 0.01 & 0.01 & 0.94 & 0.01 & 0.02 \\
\textsc{Structural} & 0.08 & 0.08 & 0.01 & 0.00 & 0.01 & 0.01 & 0.01 & 0.02 \\
\textsc{Attention} & 0.51 & 0.39 & 0.01 & 0.02 & 0.05 & 0.01 & 0.02 & 0.11 \\
\textsc{CacheBlend} & 0.19 & 0.08 & 0.09 & 0.01 & 0.01 & 0.01 & 0.01 & 0.02 \\
\textsc{Random} & 0.04 & 0.02 & 0.00 & 0.00 & 0.00 & $-$0.00 & $-$0.01 & $-$0.00 \\
\midrule
\textcolor{gray}{\textsc{CausalOracle}} & \textcolor{gray}{0.35} & \textcolor{gray}{0.42} & \textcolor{gray}{0.22} & \textcolor{gray}{0.22} & \textcolor{gray}{0.22} & \textcolor{gray}{0.09} & \textcolor{gray}{0.10} & \textcolor{gray}{0.13} \\
\textcolor{gray}{\textsc{CarrierWindow}} & \textcolor{gray}{0.61} & \textcolor{gray}{0.43} & \textcolor{gray}{0.95} & \textcolor{gray}{0.90} & \textcolor{gray}{0.89} & \textcolor{gray}{0.74} & \textcolor{gray}{0.64} & \textcolor{gray}{0.60} \\
\bottomrule
\end{tabular}
\caption{Mean MR at $K{=}32$ as the lookup table moves $d$ tokens
downstream. Sample sizes are $191/118$ (Llama), $372/359/354$ (Qwen),
and $343/216/221$ (Mistral). Bootstrap intervals are omitted. Every
half-width is at most $0.06$. The post-hoc Llama $d{=}1500$ cell is
excluded. Grey rows are diagnostics, and \textsc{CarrierWindow}
receives the true carrier location.}
\label{tab:ext}
\end{table*}

\subsection{Moving the Answer Away from the Edit}

The byte-identical $d{=}0$ tier reproduces the main results, ruling out
the two-document builder as the source of the distance effect. For
example, Qwen \textsc{EditLocal} scores $1.002$ rather than $1.007$,
and \textsc{CacheBlend} scores $0.075$ rather than $0.076$. Tests are
Holm-corrected within each four-pair cell family.

Adjacency is load-bearing. At $d \ge 250$, \textsc{EditLocal} falls to
$0.01$--$0.09$ MR and $0.00$--$0.04$ Flip. It mostly ties
\textsc{Random} on Qwen and Mistral, while Llama's remaining advantage
is only $+0.07$. \textsc{EditLocal}$@\partial$ also stops before the
carrier. \textsc{Attention} retains useful signal only on Llama,
reaching $0.39$ at $d{=}250$ and $0.84$ at $K{=}64$. It stays at or
below $0.11$ on Qwen and Mistral. The main win is therefore an adjacency
effect, not a general advantage of the selector.

Search is only part of the problem. Even with the true location,
\textsc{CarrierWindow} recovers just $0.43$--$0.74$ on Llama and
Mistral, compared with $0.89$--$0.95$ on Qwen. The probe still moves
the dependency as intended: $61$--$71\%$ of the oracle's positive
effect mass lies in the carrier block, and at most $3\%$ remains in the
edited block. Recovery from a correctly placed window is thus
model-dependent.

Llama retains only $118$ solvable items at $d{=}250$, about one third
of the pool, so its distance cells cover only solvable examples.
Distance also changes absolute position and recency, which partly
confounds the attention result.

\subsection{Transplant Recoverability Does Not Imply Recompute
Repairability}

The grey oracle ranks positions by causal effect under transplant, but
the same sets behave differently under recomputation
(Figure~\ref{fig:budget}b). On $15$ development items, transplant
recovers $0.93$--$0.99$ of the margin, while recomputation reaches only
$0.03$--$0.41$. The gap is specific to scattered sets:
\textsc{EditLocal} and \textsc{Random} differ by less than $0.02$
between operators, and all direct cells differ by less than $0.003$.

Held-out results repeat the gap on all three families. At $K{=}8$,
transplant recovers $0.918$, $0.952$, and $0.969$ on Llama, Qwen, and
Mistral, compared with $0.209$, $0.011$, and $0.008$ under
recomputation. Recomputation is worse on every held-out item, and all
$167$ recorded position sets match when replayed. Even at $K{=}32$,
recomputation reaches only $0.10$--$0.41$.

A transplanted state comes from a fully correct cache and can import
information from outside the chosen set. A recomputed state reads the
stale cache around it, so scattered positions inherit staleness. A
contiguous window instead rebuilds the forward chain in order.
Transplant recoverability therefore shows where clean information can
act, not what sparse recomputation can reconstruct. Development
diagnostics support this account: \textsc{Structural} omits content
tokens carrying most positive effect mass, while stale-query attention
ranks them too deep to fit the budget.

\subsection{Repair Should Be Unconditional}

We tested cheap repair triggers on Llama and Qwen without retention
filters ($n{=}175$ per condition). After an answer-relevant edit, the
stale cache fails on at least $98.8\%$ of items per condition, reaches
$100\%$ in both direct conditions, and never falls below $97.3\%$ in
any split. This result is limited to edits that touch the answer chain,
but within that scope there is little for a gate to separate.

Severity is no easier to predict. Ridge and random-forest models using
text, position, edit, and embedding features reach a best out-of-fold
Spearman $\rho=0.17$ ($R^2=0.05$) under grouped five-fold
cross-validation. Edit type distinguishes direct from derived cases,
but the remaining features do not support a useful severity gate.
Near-certain failure, weak predictability, and low repair cost favor
unconditional repair.

\begin{table}[t]
\centering
\small
\begin{tabular}{@{}llrrr@{}}
\toprule
Model & Ctx & Full prefill & Repair$@32$ & Speedup \\
\midrule
Llama-3.1-8B    & 4096 & 382.8\,ms & 27.8\,ms & 13.8$\times$ \\
                & 8192 & 818.1\,ms & 39.6\,ms & 20.7$\times$ \\
Qwen3-8B        & 4096 & 412.1\,ms & 31.6\,ms & 13.0$\times$ \\
                & 8192 & 884.1\,ms & 44.9\,ms & 19.7$\times$ \\
Mistral-7B      & 4096 & 374.3\,ms & 28.0\,ms & 13.4$\times$ \\
                & 8192 & 802.3\,ms & 39.8\,ms & 20.2$\times$ \\
\bottomrule
\end{tabular}
\caption{Wall-clock cost of repair (span plus $K{=}32$ downstream
positions, all layers) against a full re-prefill of the edited prefix.
Median of $20$ timed runs after $3$ warmup runs, on one RTX~5090,
bfloat16, batch $1$. Both sides
exclude the fresh-query forward, which is identical for every method.
Repair includes a defensive cache copy ($<2$\,ms), and in-place repair
is marginally faster.}
\label{tab:wallclock}
\end{table}

\subsection{What Repair Costs}

At roughly $5{,}000$ tokens, repair takes $31$--$35$\,ms versus
$458$--$502$\,ms for re-prefill, about $15\times$ faster on all three
models (Table~\ref{tab:wallclock}). The ratio reaches about $20\times$
at $8$K. Raising $K$ from $8$ to $64$ adds only $1$--$4$\,ms, making a
structural boundary such as \textsc{EditLocal}$@\partial$ inexpensive.
The token-layer proxy gives $117\times$--$234\times$ but overstates the
measured gain. Selector overhead also favors local rules, which require
no scoring forward, unlike \textsc{Attention} and \textsc{CacheBlend}.

\paragraph{Sanity checks.}
No frozen guard fired on the held-out batch. Stale and oracle outputs
reproduced byte-equal generations with margins within $10^{-3}$, and
all oracle rankings replayed without mismatches. The endpoint identity
in Eq.~\ref{eq:endpoint} passes on Llama. Qwen passes in margin
($0.997$--$1.013$) but exceeds the frozen value-tensor L2 tolerance.
A null-content control places the discrepancy within bfloat16 numerical
noise. We retain the failed label. On Mistral, three of $80$ bfloat16
smoke states changed argmax, and all three matched exactly in float32.

\section{Discussion}

\paragraph{What to deploy.}
For a direct question, refreshing the edited span completes the repair.
For an answer derived through adjacent text, recompute through the end
of the block. This rule reads only text and position, and its forward is
$13$--$21\times$ cheaper than re-prefill. A structural stopping point
such as \textsc{EditLocal}$@\partial$ avoids tuning $K$, and repair
should be unconditional because failure is nearly certain while
severity is hard to predict. This recommendation applies when the
answer dependency remains in the edited block. Beyond that boundary, a
system needs a different repair operator, not only a larger fixed
budget.

\paragraph{Coverage, not importance.}
Chunk-composition methods repair states computed from correct inputs
but missing cross-chunk attention, so large KV deviations can be useful
signals \cite{yao2025cacheblend}. An edit inside a cached document
creates a different defect: a broken dependency chain. Scattered
high-scoring positions are recomputed from stale inputs, while a
contiguous window rebuilds the path in order. Selection quality is
therefore a property of the set, not each position in isolation.
Deviation and attention scores can help when surrounding states are
sound. When they are stale, repair must cover the path that produces
the answer.

\paragraph{Localization evidence overstates what repair can do.}
Cache-editing work can rank positions by patching clean states into a
corrupted run \cite{li2026MTN}, but that ranking need not transfer to
recomputation. Transplant imports states produced in a correct context,
including information from outside the chosen set. Recomputation must
rebuild those states from the stale cache. The operator gap shows that
a rule intended for recomputation must be tested under recomputation.
Transplant locates where clean information can influence the output,
but it is not an upper bound on deployable repair.

\paragraph{Where repair still fails.}
Once dependent text moves $250$ tokens away, \textsc{EditLocal}
collapses and only stale-query attention on Llama provides a useful
deployable alternative. Search is not the whole problem:
\textsc{CarrierWindow} knows the carrier location yet remains far below
full recovery on Llama and Mistral. A second pass or a window spanning
the edit-to-carrier path may restore the missing inputs, but neither is
tested here. Qwen's stronger result also shows that the boundary is
model-dependent.

\paragraph{Limitations.}
Edits are single, contiguous, and length-preserving, and the synthetic
blocks make every edit answer-relevant. The failure rate therefore does
not extend to arbitrary edits, and the two-value metrics do not measure
long free-form answers. Llama retains $34$ of $75$ held-out derived
items and about one third at $d{=}250$, so those cells cover only
solvable examples, and distance also changes recency. The transplant panel
uses $15$ development items without Mistral, although the operator gap
is also measured on held-out data. \textsc{EditLocal}$@\partial$ was
inferred from unblinded curves rather than tested as a registered arm.
The models are dense and near $8$B, and timing uses one GPU at batch
$1$ while excluding the identical query forward. Multi-edit and
length-changing updates, larger or sparse models, and production
serving remain open.

\section{Conclusion}

We framed in-place repair of a stale KV cache as budgeted recomputation
and compared training-free selection policies at a matched budget on
paired factual edits across three model families. Recomputing a
contiguous window from the edit to the end of its block restores
post-edit behavior almost completely wherever the dependent text is
adjacent, beats every signal-based rule by a wide margin, and runs
$13$--$21\times$ faster than a re-prefill. The result carries a
mechanism and a boundary. Repair works by rebuilding a forward
dependency chain, so position sets that look sufficient under transplant
fail under recomputation, and the same window fails once the chain grows
long. Edit-local recomputation is the baseline that any future policy
for this problem has to beat.


\bigskip

\bibliography{aaai2027}


\end{document}